%% file: paper.tex
\documentclass[11pt,a4paper]{article}
\usepackage[a4paper, 
            left=1.25in,
            right=1.25in,
            top=1in,
            bottom=1in]{geometry}
\usepackage{color}
\usepackage[colorlinks = true,
            linkcolor = blue,
            urlcolor  = blue,
            citecolor = blue]{hyperref}
\usepackage{amsmath, amsfonts, amsthm, amssymb}
\usepackage{bm} 
\usepackage[super]{nth}
\usepackage{verbatim}
\usepackage{stmaryrd}
\usepackage{float}
\usepackage{fancyhdr}
\usepackage{color}
\usepackage{graphicx}
\usepackage{subfig}
\usepackage{wrapfig}
\usepackage[capitalise]{cleveref}
\usepackage{booktabs}
\usepackage{mathtools}
\usepackage{natbib}
\usepackage{nicefrac}
\usepackage{shuffle}
\usepackage{authblk}

\input{other/math_commands}

\usepackage[parfill]{parskip}

\usepackage{multicol, caption}
\theoremstyle{definition}

\newtheorem{example}{Example}[section]

\crefname{lemma}{lemma}{lemmas}
\Crefname{lemma}{Lemma}{Lemmas}
\crefname{theorem}{theorem}{theorems}
\Crefname{theorem}{Theorem}{Theorems}
\crefname{corollary}{corollary}{corollaries}
\Crefname{corollary}{Corollary}{Corollaries}
\Crefname{proposition}{proposition}{propositions}
\Crefname{proposition}{Proposition}{Propositions}
\crefname{remark}{remark}{remarks}
\Crefname{remark}{Remark}{Remarks}
\crefname{definition}{definition}{definitions}
\Crefname{definition}{Definition}{Definitions}
\crefname{example}{example}{examples}
\Crefname{example}{Example}{Examples}
\crefname{assumption}{assumption}{assumptions}
\Crefname{assumption}{Assumption}{Assumptions}
\crefname{axiom}{axiom}{axioms}
\Crefname{axiom}{Axiom}{Axioms}

\title{New directions in the applications of rough path theory}

\author[1,2,3]{Adeline Fermanian}
\author[4,6]{Terry Lyons}
\author[4,6]{James Morrill}
\author[5,6]{Cristopher Salvi}

\affil[1]{MINES ParisTech, PSL Research University, CBIO, F-75006 Paris, France}
\affil[2]{Institut Curie, PSL Research University, F-75005 Paris, France}
\affil[3]{INSERM, U900, F-75005 Paris, France}
\affil[4]{Department of Mathematics, University of Oxford}
\affil[5]{Department of Mathematics, Imperial College London}
\affil[6]{The Alan Turing Institute}

\date{}

\begin{document}

\maketitle

\pagenumbering{arabic}

\begin{multicols}{2}

\input{sections/introduction}
\input{sections/streams}

\input{sections/ncdes}

\input{sections/rnns}
\input{sections/kernels}

\input{sections/conclusion}

\bibliography{references}
\bibliographystyle{apalike} 

\end{multicols}
\end{document}

%% file: other/math_commands.tex
\newcommand{\mb}[1]{\mathbf{#1}}

\def\reals{{\mathbb{R}}}

\def\to{{\rightarrow}}
\def\dd{\mathrm{d}}

\def\logsig{\mathrm{LogSig}}
\def\sig{\mathrm{Sig}}

%% file: sections/introduction.tex
\section*{Abstract}

This article provides a concise overview of some of the recent advances in the application of rough path theory to machine learning. Controlled differential equations (CDEs) are discussed as the key mathematical model to describe the interaction of a stream with a physical control system. A collection of iterated integrals known as the signature naturally arises in the description of the response produced by such interactions. The signature comes equipped with a variety of powerful properties rendering it an ideal feature map for streamed data. We summarise recent advances in the symbiosis between deep learning and CDEs, studying the link with RNNs and culminating with the Neural CDE model. We concluded with a discussion on signature kernel methods.

\section{Introduction}


Complex and messy data streams (rough systems) are everywhere in our lives and come in profoundly different forms. They can take the form of a conversation, an electronic health record of a patient, the gravitational interaction of planets in our solar system, or the complex flow of traffic in a city. It is a profound and unfinished challenge to effectively model such streams, capture their ``meaning", and understand their interactions in a way that transcends the individual context. 

Rough path theory is an abstract mathematical tool-set that allows the modelling and analysis of evolving systems across a broad spectrum of use cases. It extends Newton's and It\^{o}'s calculus to model the interactions of complex and messy data streams. It has been influential within mathematics, in particular in stochastic analysis, as demonstrated by Hairer's 2014 Fields Medal citation notes: \textquotedblleft Building on the rough-path approach\ldots Hairer was thus able to give, for the first time, a rigorous intrinsic meaning to many Stochastic Partial Differential Equations (SPDEs) arising in physics.\textquotedblright. Like the mathematics of the number system, and the calculus of differential equations, the theory of rough paths exposes an aspect of the modelling of these streams that is separate from the context of interest (conversations, health records, solar system...).

A key insight in classical reservoir computing is that the trajectory of a dynamical system can be effectively described through its interaction with a 'black-box' non-linear system capable of storing information. In rough path theory, the dynamical system is replaced by a generic stream, the rough path, and the black-box by a simple and carefully chosen 'white-box' system, which fully describes the unparameterised stream \citep{hambly2010uniqueness}. The outcomes of these interactions build a \textquotedblleft structured description of the rough system\textquotedblright on a macroscopic scale. This mathematical description in terms of \textquotedblleft effects\textquotedblright is rich enough to characterise internal interactions and create a new calculus, without needing to directly determine its microscopic structure.

A traditional approach for handling time series is to record their values at regular intervals. Simple real-world examples demonstrate that this approach to describe streamed data is neither scalable nor practical. For example, consider the task of predicting the total CO2 emissions of a household over a month given the time series of their electricity consumption. The signals would need to be sampled significantly more often than 50 times a second to determine the power consumption. Rough path theory represents a radical change in the way one thinks mathematically about streamed data. It  summarises   the stream through a canonical sequence of coefficients, called the signature. The first three terms in the signature of the electrical stream provide sufficient information to compute the power consumption.




The signature of a stream can be viewed as a generating function economically recording the effects of this stream on generic nonlinear systems. The geometer K.T. Chen used
this approach to understand multidimensional paths in the 1950s \citep{chen1954iterated}.

There are several substantial differences between the signature and classical signal processing transforms such as Fourier series and wavelets. One of the most significant is that Fourier series and wavelets are linear transformation and so treat the channels in a multimodal stream independently. As a result these transforms cannot capture information related to order of events occurring across the different channels. For example, regularly seeing the sequence: phone call, trade, price movement in the stream of office data monitoring a trader might be an indication of insider trading. Such occurrences
are straightforward to detect via a linear regression composed with the signature transform.
Modelling this signal using Fourier series or wavelets could be much more expensive: linearity
of these transforms imply that each channel must be resolved accurately enough to see the order of
events. 

The signature has the powerful property of being invariant to time-reparameterization of the underlying stream, which can have a dramatic effect on the complexity of certain challenges. In other words, the signature is a non-linear filter which removes the dependence on the sampling rate, while retaining all other information in the stream (in particular the order of events).

To sum up, rough path theory uses a top-down approach to summarise and digitise a stream. It uses a new feature set (signature) taken on longer timescales of interest, still capturing sufficient information to predict effects of the stream (the P\&Ls of different investment strategies for example). These summaries can be orders of magnitude more parsimonious without losing their effectiveness. This top-down methodology is in strong contrast to the bottom-up, finer and finer steps approach of classical integration and calculus.

Combined with state-of-the-art machine learning algorithms, the rough path approach for describing data streams has proved to be powerful for many real-world problems \cite{kidger2019deep}. For example, Hans Buehler says, in his ``2022 quant of the year award'' article: ``Signatures, thanks to their ability to converge to a stable result with a relatively small training dataset, proved perfectly suited to the task.'' 

This article is organised as follows. Section \ref{sec:streams} introduces controlled differential equations (CDEs) and signatures. Section \ref{sec:nn_cdes} discusses how deep learning and rough path theory can be combined to learn the ``physics'' of a controlled system from observational data. Section \ref{sec:kernels} discusses signature kernel methods.

%% file: sections/streams.tex
\section{CDEs and signatures}\label{sec:streams}


For this article a stream is a multidimensional evolving state that has effects. That is to say it can interact, over time intervals, with systems in transformative and measurable ways. 

To match with standard mathematical approaches, the reader might restrict attention to streams with an evolving state that can be recorded adequately as a smoothly sampled time series, denoted by $\boldsymbol{X}:[s,t) \rightarrow V$, where $V$ is a vector space, but they should allow that the sampling timescale may be orders of magnitude finer than the scales on which the effects they are interested in occur. Financial markets (micro-secs, days), astronomy (GHz, ms), speech (4kHz, .5 sec) and the power consumption of electric appliances (50Hz, minutes) all provide examples of this separation of timescales. In most cases, $V$ is chosen to be $\reals^d$, $d \geq 1$.

\subsection{CDEs}
The mathematical modelling of streams via controlled differential
equations has a long history. Already, in Newton's work and language there were\
systems that evolved, with position (fluent) and direction (fluxion). They
interacted and there was a notion of a control (relate or dependent
variable; and correlate or independent variable) and he solved for the
evolution of the fluents.\footnote{%
Newton, Isaac, Methodus fluxionum et serierum infinitarum, 1671; The Method
of Fluxions and Infinite Series; with Its Application to the Geometry of
Curve-lines... Translated from the Author's Latin Original Not Yet Made
Public. To which is Subjoin'd a Perpetual Comment Upon the Whole Work... by
J. Colson. 1736.}


The language of controlled differential equations has matured. Consider first a conventional ordinary differential equation governed by a vector field $f$
\begin{eqnarray*}
\dd \boldsymbol{Y}_t = f\left( \boldsymbol{Y}_t\right) dt
\end{eqnarray*}
Such differential equation defines a flow; one can start the system at any point (the initial condition), but there are no other degrees of freedom for the dynamics. A controlled differential equation (CDE) is a differential equation where the dynamics, which are governed by $f$, can change according to an external stream of multimodal information $\boldsymbol X$ (called the control):
\begin{eqnarray}\label{eq:cde}
    \dd \boldsymbol{Y}_t = \boldsymbol{f}\left( \boldsymbol{Y}_t\right) \dd \boldsymbol{X}_t
\end{eqnarray}
Here, $\dd \boldsymbol{Y}_t$ represents the infinitesimal response of the system, $\dd \boldsymbol{X}_t$ the infinitesimal control, and $\boldsymbol{f}$ is a linear function mapping $\dd \boldsymbol{X}_t$ to vector fields on the $\boldsymbol{Y}$ space. CDEs also define a flow however there are infinitely many degrees of freedom.

\subsubsection*{Other Forms of CDE}
Non-autonomous equations like 
\begin{equation*}
\dd \boldsymbol{Y}_{u}=\boldsymbol{f}\left( \boldsymbol{Y}_{u},\boldsymbol{u}%
\right) \dd \boldsymbol{X}_{u}+\boldsymbol{g}\left( \boldsymbol{Y}_{u},%
\boldsymbol{u}\right) \dd \boldsymbol{u}
\end{equation*}%
look more general but they are not. We are already working with
multidimensional data, and by augmenting the state and control variables 
\begin{equation*}
\boldsymbol{Y}_{u}\rightarrow \left( \boldsymbol{Y}_{u},\boldsymbol{u}%
\right) ,\ \ \boldsymbol{X}_{u}\rightarrow \left( \boldsymbol{X}_{u},%
\boldsymbol{u}\right) 
\end{equation*}%
and modifying the physics%
\begin{eqnarray*}
\dd \boldsymbol{Y}_{u} &=&\boldsymbol{f}\left( \boldsymbol{Y}_{u},\boldsymbol{u}%
\right) \dd \boldsymbol{X}_{u}+\boldsymbol{g}\left( \boldsymbol{Y}_{u},%
\boldsymbol{u}\right) \dd \boldsymbol{u} \\
\dd \boldsymbol{u} &\boldsymbol{=}&\dd \boldsymbol{u}
\end{eqnarray*}%
we may recast non-autonomous control problems as autonomous problems in an
enhanced state space. We may do the same with other more complex
intertwinings like $\dd \boldsymbol{Y}_{u}=\boldsymbol{f}\left( \boldsymbol{Y}%
_{u},\boldsymbol{X}_{u}\right) \dd \boldsymbol{X}_{u}$. In conclusion, it suffices to
understand the fundamental equation 
\begin{equation*}
\dd \boldsymbol{Y}_{u}=\boldsymbol{f}\left( \boldsymbol{Y}_{u}\right) \dd \boldsymbol{X}_{u}.  \label{eqn1:the fundamental controlled equation}
\end{equation*}

\subsection{Describing a Stream with Signatures} \label{sec:signatures}

The simplest non-trivial CDE is probably the exponential, which for any $u \in [s, t)$, may be written informally as
\begin{equation}
\dd \boldsymbol{Y}_{u}=\boldsymbol{Y}_{u} \cdot \dd \boldsymbol{X}_{u},\quad 
\boldsymbol{Y}_{s}=1,  \label{eqn1:the signature equation}
\end{equation}%
where $\cdot$ denotes a multiplication to be defined later. Here, the evolution of $\boldsymbol{Y}$ exposes the effect of the stream $\boldsymbol{X}$ on the simplest non-linearity---multiplication. The measured response $\boldsymbol{Y}_{t}$ of this controlled system to the stream $\boldsymbol{X}_{u}$ over the interval $u\in \lbrack s,t)$ yields precise information about the stream over the interval. The geometer KT Chen used
this approach to understanding multimodal paths in the 1950s. The \emph{signature} $\sig_{s,t}(\boldsymbol{X})$ of the stream $\boldsymbol{X}$ on $[s,t)$ is defined as the response $\sig_{s,t}(\boldsymbol{X}):=\boldsymbol{Y}_{t}$. In other words, the signature $\sig_{s,t}(\boldsymbol{X})$ summarises the small-scale changes of $\boldsymbol{X}$ on $[s,t)$.

\subsubsection*{Tensors}

The equation \eqref{eqn1:the signature equation} implicitly requires that $%
\boldsymbol{Y}$ takes values in a space where one can multiply any element $%
\boldsymbol{Y}$ by any element $\boldsymbol{x}$ of $V$.
One gets the fullest description of $\boldsymbol{X}$ from the response $%
\boldsymbol{Y}$ if the algebra of the target space for the differential
equation imposes no other relations. The space generated by $V$ and a free
non-commutative multiplication operator is known as the tensor algebra. The
full signature is an element of this space, it is a sequence of tensors over
V; one of each degree. For those interested in details, we explain.

To make the exposition concrete, let $\boldsymbol{v_1},\boldsymbol{v_2},\ldots ,\boldsymbol{v_d} $ be a finite collection of letters (which we think of as an alphabet $A$) and suppose these letters are members of and form a basis for the linear space $V$. That is to say, every element $\boldsymbol{x}\in V$ is a unique linear combination $\sum_{i} x_i \boldsymbol{v_i} $ of letters in $A$. 

Now define $A^{\ast }$ to be the set of all words, including the empty word, whose
letters are drawn from $\boldsymbol{v_1},\boldsymbol{v_2},\ldots ,\boldsymbol{v_d}$. If $\boldsymbol{\varepsilon } $ is the empty word then 
\begin{align*}
A^{\ast }=\big\{& \boldsymbol{\varepsilon} ,\boldsymbol{v_1},\boldsymbol{v_2},\ldots, \boldsymbol{v_d}, \\
&\boldsymbol{v_1 v_1}, \boldsymbol{v_1 v_2},\ldots ,\boldsymbol{v_1v_d}, \\
&\boldsymbol{v_2v_1},\boldsymbol{v_2 v_2},\ldots \big\},
\end{align*}
and $T((V))$ is the vector space of all formal linear combinations of words
in $A^{\ast }$.  One can split this sum by the length of the word; one can write 
\begin{equation*}
\boldsymbol{y}=\sum_{\boldsymbol{w}\in A^*}y_{\boldsymbol w}\boldsymbol{w} \in T((V))
\end{equation*}
as
\begin{align*}
\boldsymbol{y}&= \overset{\text{degree 0}}{\vphantom{\sum_{A}}y_{\boldsymbol{\varepsilon }}\boldsymbol{\varepsilon }} + \overset{\text{degree 1}}{\sum_{
\boldsymbol{v_i}\in A}y_{\boldsymbol{v_i} }\boldsymbol{v_i}} \\
& \quad + \overset{\text{degree 2}}{\sum_{\boldsymbol{v_i,v_j}\in A}y_{\boldsymbol{v_iv_j}}%
 \boldsymbol{v_i v_j}} \\
 & \quad + \overset{\text{degree 3}}{\sum_{\boldsymbol{v_i,v_j,v_k}\in A}y_{\boldsymbol{v_iv_jv_k}} \boldsymbol{v_i v_j v_k}} + \cdots .
\end{align*}
The word of length zero yields a constant, the words of length one yield a
vector in $V$ while the words of length $n$ yield a tensor of degree $n$
over $V$. An element of $T((V))$ is an infinite sequence of tensors,
starting with a constant or 0 degree tensor, then a vector in $V$ or a $1$%
-tensor, then a matrix or a $2$-tensor, $3$-tensor, and so on.

The space $T((V))$ is an algebra; multiply words by concatenation 
\begin{align*}
    &( \boldsymbol{v_{i_1}}\ldots
\boldsymbol{v_{i_k}}) \left( \boldsymbol{v_{j_1}}\ldots \boldsymbol{v_{j_\ell}}\right) \\ 
& \quad = \boldsymbol{v_{i_1}}\ldots \boldsymbol{v_{i_k}}\boldsymbol{v_{j_1}}\ldots \boldsymbol{v_{j_\ell}},
\end{align*}

and extend their product to sums of words via the distributive
property. Multiplication is denoted by $\otimes $ for generic elements of $%
T\left( \left( V\right) \right) $ and sometimes for  concatenation of elements in $A$ for notational simplicity. The empty word $\boldsymbol{\varepsilon}$ is the unit and is  written as $\boldsymbol{1}$.

\emph{The signature of a path is an element of $T((V))$. It is a tensor of
every degree, or a coefficient associated to every word. One readily checks
that in the case where on some very fine scale }$\boldsymbol{X}$ is a
smooth path, and classical calculus applies, that the $n$-tensor term $%
\sig_{s,t}^{n}(\boldsymbol{X})$ in the signature is described recursively in terms
of the $(n-1)$-tensor term $\sig_{s,t}^{n-1}(\boldsymbol{X})$ without reference to the
higher order terms and
\begin{equation}
\sig_{s,t}^{n}(\boldsymbol{X})=\int_{u=s}^{t}\sig_{s,u}^{n-1}(\boldsymbol{X})\otimes \dd \boldsymbol{X}_{u}  \label{eqn3:iterated_integral_signature}
\end{equation}%
solves the differential equation (\ref{eqn1:the signature equation}) at
every tensor degree.

Back substitution in the formulation \eqref{eqn3:iterated_integral_signature} for the solution quickly reveals a second more popular expression for the
signature $\sig_{s,t}( \boldsymbol{X})$ in terms of iterated integral expansions.%
\begin{align*}
 \boldsymbol{1} & +\int_{s<u<t}\dd
\boldsymbol{X}_{u} \\
&  +\int_{s<u_{1}<u_{2}<t}\dd \boldsymbol{X}_{u_{1}} \otimes \dd \boldsymbol{X}_{u_{2}} + \cdots .
\end{align*}%
Conceptually, the approach via effects offers a way to compute
the signature of a stream that, if non-linear sensors are available, is far
less computationally intensive than computing iterated integrals. The approach via
effects also allows one to deal with streams that have jumps in a much more
principled way.

\subsubsection*{The truncated signature}


Real numbers in a computer are traditionally truncated and represented using some version of IEEE Floating-Point Arithmetic. The tensor algebra T((V)) is managed similarly. By sending every word with more than $n$ letters, every tensor of degree greater than $n$, to zero, we get the truncated tensor algebra $T^{(n)}(V)$.

An element of this truncated tensor algebra is a sequence of tensors of degree up to and including degree $n$. Multiplication and addition remain the same except that words of greater degree than $n$ that arise are set to zero and forgotten. 

The signature of the stream $\boldsymbol{X}_u,\ u\in [s,t)$ has tensor components of every positive integer degree $n$ and the number of degrees of freedom in a component grows exponentially with this degree! The complexity of $\boldsymbol{X}_u,\ u\in [s,t]$ determines the degree of the truncated signature needed to effectively describe $\boldsymbol{X}$. A path whose $n$ increments are drawn from an alphabet with $d$ letters already admits $d^n$ possibilities. 

If $V$ is finite dimensional and $d=\text{dim}(V)$ then the truncated algebra $T^{(n)}(V)$ has dimension $(d^{n+1} - 1)/(d-1)$; if $d>1$ the tensor of degree $n$ constitutes more than half the dimension of the entire truncated feature set. Because of this, one has the slightly surprising result that the computational complexity of multiplying two elements in the truncated tensor algebra is only logarithmically more than the cost of touching the data. This is in contrast to matrix multiplication which is much more expensive than the cost of touching the matrix. Numerical computation of signatures with well written code is primarily limited by memory bandwidth and is viable in quite high dimensions despite the exponential growth of the feature set.

The trick to using signatures to describe streams is to understand that they are very effective at describing oscillations that happen on a short time scale relative to matters of interest. Describing a stream effectively over a short interval may only require a few terms in the  signature and still be incredibly economical and accurate. However, shrinking the time intervals beyond this point becomes expensive because the data associated to lots of short time intervals is duplicate and irrelevant. This is the conundrum. 
The sweet spot is typically intermediate. Describing the stream over moderate length intervals using moderate depth truncated signatures can lead to dramatic efficiencies. 

\begin{example}[Electricity bill]
A good every day example is in the electricity bill. Electricity is delivered as a 2 dimensional time varying signal of voltage $V_t$ and changing charge $Q_t$. Both alternate, changing sign 50-60 times a second. 

From the perspective of the electric motor this short term description might be vital, but for many purposes the signal is adequately described by a single figure once a month - the total power consumed. In fact this total power consumption is the first significant term in the signature of the $(V_t,Q_t)$ path (it is a second order term - precisely the area and needs to be reported once in the month). This single nonlinear 2-truncated signature descriptor of the electric signal over the month describes the 129.8M oscillations well enough to allow prediction of CO$_2$ output etc. 

It is an interesting open exercise to identify the most parsimonious data store of truncated signatures over intermediate time intervals that is sufficiently information rich to allow identification of devices as they are switched on and off. The raw data is huge but not directly informative.
\end{example}

\subsubsection*{Signatures bring algorithmic improvement}

Signatures and controlled differential equations, when used to generate feature sets, remove parameterisation from a data stream, which can have a
dramatic effect on the complexity of certain challenges. We illustrate this with
a simple example which we call the insider trader problem. We first explain
the problem in the language of classical time series, and bucket data into
discrete time points.

\begin{example}[Insider trader problem]
Investigators have the records of several traders suspected of insider
trading, they would like to train a naive machine learnt function that they
could apply to other data to identify other suspected insider trader. Each
trader had associated with them a time series $\boldsymbol{x}:=\left\{ \left(
x_{i},y_{i},z_{i}\right) ,i=0,1,\ldots \right\} $ where $i$ represents the
minutes of the working day. Most times $x_{i},y_{i},z_{i}$ are all zero, but
(in this thought example) once a day the trader makes a short phone call; $%
x_{i}=1$ at the minute $i$ that the trader initiates her call. Once a day
the trader makes a trade in some asset; $y_{i}=1$ at the minute $i$ that the
trader initiates her trade. Once a day the
trader's asset has a value change; $z_{i}=1$ at the minute $i$ that the
asset price moves. There are $480$ minutes in a working day. The time series
has $3$ dimensions at each time or $1440$ dimensional in total. 

Suspect insider traders make the phone call, make the trade, and see the
price movement \emph{in that order}; any other order for these interactions
is not suspicious. Now consider a function $f$ that is $1$ if $i\leq j\leq k$
and zero otherwise and which identifies the suspects. A simple cubic
polynomial will do the job:%
\begin{equation*}
f\left( \boldsymbol{x}\right) :=\sum_{0\leq i<j<k<480}x_{i}y_{j}z_{k}
\end{equation*}%
but one immediately sees a problem; there are a huge number of ways that $%
i<j<k$ and even bigger complimentary set of non-suspicious triples. Training
to find the function $f$ cannot begin to get things correct unless it is
trained on most of these time series instances. $\left( 480\right) ^{3}$of
them. To see the pattern needs a huge amount of data and the situation gets
worse by several orders of magnitude if one refines the time scales.
Learning order information in multi-modal time series is extraordinarily
expensive.

However, adopting a signature approach transforms the situation. Let us
replace the discrete time series by a database of event tick data. The
database has three entries, $I$ the time of the phone call; $J$ the time of
the transaction; and $K$ the time of the asset price change. It is easy to
construct signatures from time stamped categorical tick data and the
complexity is $NC^{d}$ where $N$ is the number of ticks, $C$ is the number
of categories, and $d$ is the truncation degree of the signature. In our
case the number of ticks and categories is both $3$. Now, $f$ only depends
on the order and not the timing and so one knows that it can be realised as
a linear functional of the signature. It is actually a coefficient of the
signature truncated at degree $3$.


So we have replaced our $1440$ dimensional time series representation of our
stream, to a $3^{4}$ or $81$ dimensional one and the function of interest is
a linear functional on this signature feature set. We have quite
dramatically reduced the amount of data required to learn the function. 
\end{example}


\subsubsection*{The log-signature}

The signature is in some sense the exponential of the stream over a time interval---see Equation \eqref{eqn1:the signature equation}. It is the universal nonlinear response to the stream and any such response is well approximated by a linear projection of the signature. 

The log-signature summarises the stream in a somewhat different manner but plays an equally valuable role. It is a more concise descriptor of the stream over a time interval. It is a vector that is as easy to compute as the signature but has no redundancy. Polynomial functions in the log-signature have universality instead of the linear functions of the signature.

Its usefulness comes from the ease with which it can predict the effects of $\boldsymbol{X}$ on a a given non-linear system.  Consider again the equation 
\begin{eqnarray*}
\dd \boldsymbol{Y}_{u} &=&\sum_{i}f^{i}\left( \boldsymbol{Y}_{u}\right)
\dd X_{u}^{i} 
\end{eqnarray*}%
capturing the impact of $\boldsymbol{X}$ over a short interval of time $[s,t)$. Those interested in numerical solution of stochastic differential equations have for many years found it convenient to replace this system  with another \emph{time invariant} differential equation 
\begin{align*}
\dd\boldsymbol{y}_{v} = \boldsymbol{F}\left( \boldsymbol{y}_{v}\right)
\dd v, \quad v\in[0,1] 
\end{align*}
that captures the same net effect over $[0,1]$ to a high order of approximation as the controlled differential equation does over $[s,t)$. The choice of the differential equation (or vector field $\boldsymbol{F}$) depends on the stream over the interval $[0,t]$. But its solution can be delegated to classical ODE solvers. It is known as the log-ode method. It is usually one of the most efficient approaches to solving SDEs numerically. 

The choice of $\boldsymbol{F}$ is totally canonical and  is determined by the $f^i$ and is a linear combination of lie brackets of the $f^{i}$ determined by the \emph{log-signature} of $\boldsymbol{X}$ over the interval $[s,t)$. It is beyond this short note to explain this in detail, but should not be unfamiliar to those interested in control theory with knowledge of the the works of Brockett and Sussman, etc.

The log-signature is usually calculated by applying the power series for $\log(1+x)$ to the signature with the leading one removed. The log-signature is a linear combination of formal lie brackets $[\boldsymbol{v}_{i_1}, \ldots, [\boldsymbol{v}_{i_j},\boldsymbol{v}_{i_k}], \ldots,\boldsymbol{v}_{i_\ell}]$, which are easily translated to vector fields by substituting $f^i$ for $\boldsymbol{v}_{i}$. 
The first term in the log-signature is the increment of the stream in each coordinate, the second is the pairwise area. Higher order terms are easy to compute but require a choice of basis for the free Lie subalgebra of the Tensor algebra. There is a strong mathematical theory of these objects going back (among many others) to Hall, Schutzenberger, Ree, and Bourbaki \citep{ reutenauer2003free}. The software packages iisignature(PyPi) and esig(PyPi) both offer such calculations and the library libalgebra(C++, github) offers a broad range of related tools.

\begin{figure*}
    \centering
    \includegraphics[width=\textwidth]{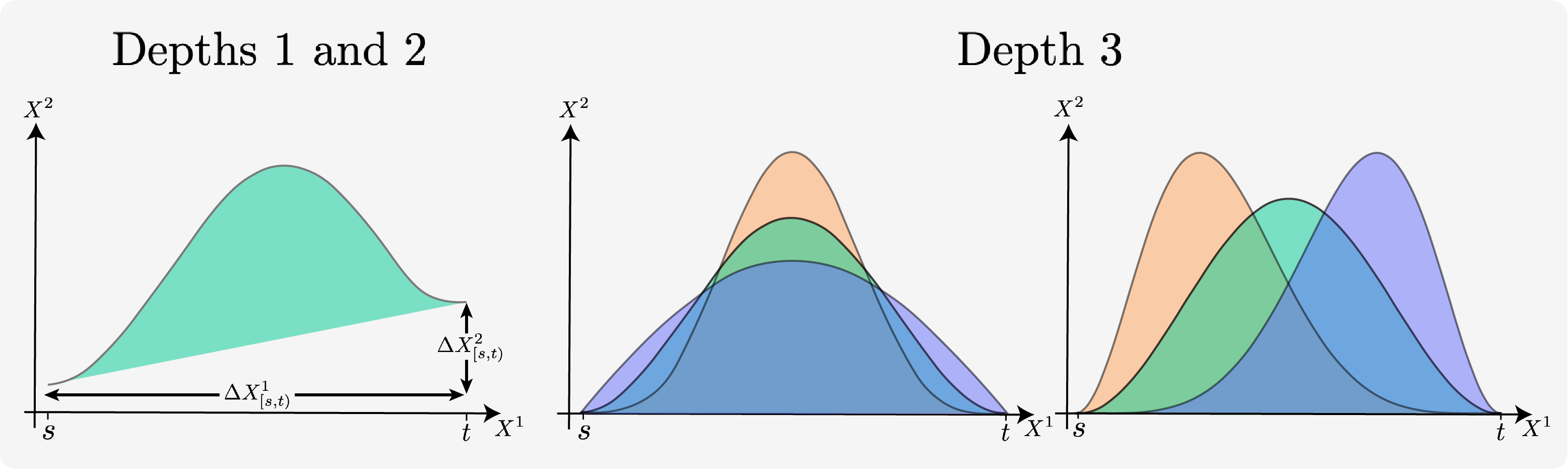}
    \caption{Graphical depiction of the depth 1, 2, and 3 terms of the log-signature for a two-dimensional path $(X^1, X^2)$. \textbf{Left:} The first two depths that includes the change of the variables over the interval and the L\'evy area (shaded). \textbf{Middle and right:} Each plot only varies in a single depth 3 term with other terms being fixed. The middle plot sees varying area density along the $X^2$ axis, and the right along the $X^1$ axis.}
    \label{fig:interpretability}
\end{figure*}

\subsubsection*{Geometric interpretation}

Aside from being grounded in strong mathematical theory, the signature of a path also has geometric meaning. This is of particular interest when learning functions on data, since use of the signature can result in models that are interpretable and explainable. 

The left most plot in Figure \ref{fig:interpretability} contains the visualisations of both the depth 1 and 2 terms of the log-signature. The depth 1 terms are simply the change of the features over the interval; this can be seen directly via
\begin{align*}
    \logsig^1_{s,t}(\boldsymbol{X}) 
    &= \int_{s < u < t} \dd \boldsymbol{X}_u \\
    &= \boldsymbol{X}_t - \boldsymbol{X}_s \\
    &= \Delta \mb{X}_{[s, t)}.
\end{align*}

The depth 2 term is the signed area between the path and the chord joining the endpoints. This is also known as the \textit{L\'evy area} of the path; further examples can be found in \citet{chevyrev2016primer}. 

There exist two more terms at depth 3, however, they are more difficult to visualize directly. However, consider the middle and right plots in \cref{fig:interpretability}. In each plot we draw three curves that differ only in their log-signature values at one of their depth 3 terms, with the same depth 1 and 2 values. The plot in the middle shows how one of the terms at depth 3 captures the distribution of the area along the $X^2$ axis. The right most plot demonstrates that the final term captures the extent to which the area has been loaded along the $X^{1}$ axis. 

\begin{example}[Interpretability of breathing data]
    To demonstrate how these features may be interpretable in the context of real data, consider a situation where we are seeking to predict stress from a persons respiratory data (a  problem where signatures have proved very successful). Consider each of the curves in \cref{fig:interpretability} to represent a single breath, with $X^1$ being time and $X^2$ the percentage expansion of a band located around the wearers chest that oscillates with breathing \citep[see][]{WESAD2018}.
    
    The values in the signature to depth 1 represent the total time of the breath ($\Delta X^1)$ and the difference in chest expansion from the start and end of the breath ($\Delta X^2$). The depth 2 term (the shaded area) approximates the volume of the inhalation. Finally, the depth 3 term can be seen to represent the sharpness of the inhale (\ref{fig:interpretability}, middle) and the extent to which the breath was a sharp inhale followed by a long exhale, or vice versa.
    (\ref{fig:interpretability}, right).
\end{example}


Now that we have presented the main mathematical tools in applied rough path theory, we describe how these objects are applied in machine learning.

%% file: sections/ncdes.tex
\section{Deep learning and CDEs} \label{sec:nn_cdes}

Over the last few decades, neural networks have emerged as a spectacular new way of learning functional relationships from data. These functional relationships span a wide range of tasks such as computer vision or language generation, as well as based around the interpretation of streamed data. 

One of the exciting developments in this domain over the last couple of years has been the realisation that neural networks and differential equations are not competitors for describing functions on streams. Instead, there exists substantial synergy between the two models. This relationship was first stressed in the context of neural ordinary differential equations (neural ODEs), where neural networks are used as high capacity function approximators to learn ODEs \citep{chen2018neural}, and subsequently extended to other types of equations such as CDEs \citep{kidger2020neural, morrill2021neural}.


In the next section, we outline how the mathematical theory introduced in \cref{sec:streams} can be used in tandem with neural networks to model functions on irregularly-sampled and partially observed time series, in the context of neural CDEs.





\subsection{\fontsize{14}{14}Neural CDEs} \label{sec:ncdes}

\begin{figure*}
    \centering
    \includegraphics[width=\linewidth]{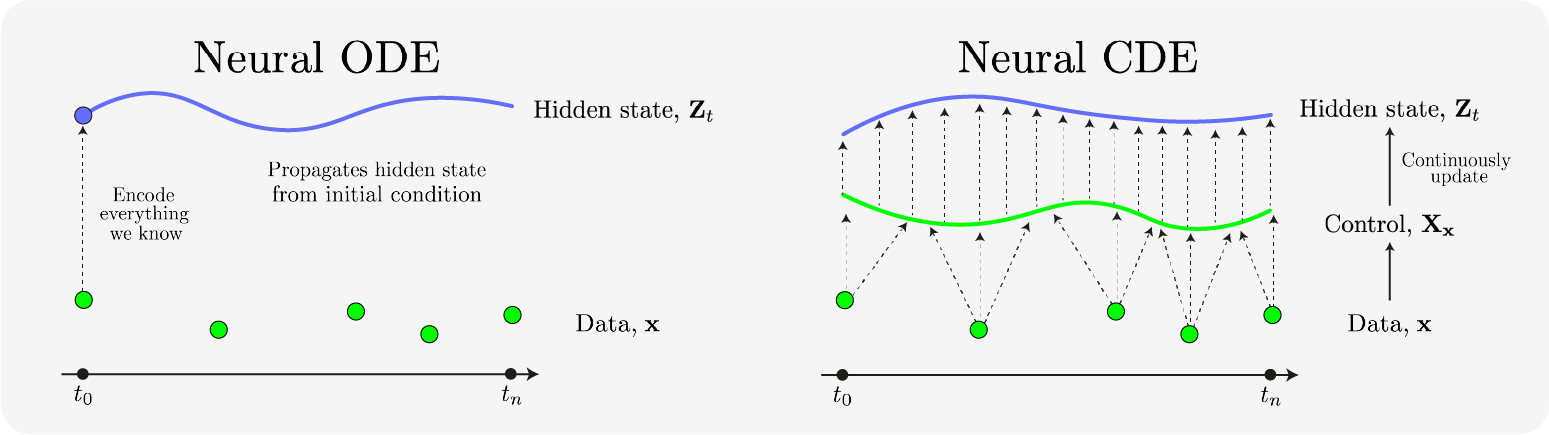}
    \caption{Visual comparison of a Neural ODE vs a Neural CDE. \textbf{Left:} The Neural ODE encodes all information available to us at $t_0$ and propagates the hidden state forward in time, with no alterations based on future data. \textbf{Right:} The Neural CDE model has a mechanism to incorporate new data in continuous time.}
    \label{fig:node_vs_ncde}
\end{figure*}

We assume that we observe some data with their time-stamps: we are given a sequence 
\begin{equation*}\label{eq:def_data_sequence}
    \boldsymbol{x} = \big((t_1, x_1), \ldots, (t_T, x_T) \big),
\end{equation*}
where $t_i \in \reals$ are time-stamps such that $t_i < t_{i+1}$ and $x_i \in \reals^{d-1}$ are the features. The data may be both irregular (variable gaps between the $t_i$'s) and partially observed (not all features need to be measured at each $t_i$). 

To model the discrete data $\boldsymbol{x}$ using a CDE we must first formulate a continuous time embedding of the data, since the CDE evolves under the influence of a continuous time control $\boldsymbol{X}$. A sensible method is to use an interpolation of the data, typically such that $(t_j, \boldsymbol{x}_j) = \boldsymbol{X}_{t_j}$, $1 \leq j \leq T$; an in depth description of how one might do this is given by \citet{morrill2021neuralonline}. From now on, we denote by $\boldsymbol{X}$ the embedding of $\boldsymbol{x}$ as a continuous-time process.

The general idea of a neural CDE is to model a target hidden state $\boldsymbol{Z}_t$ which lives in $\reals^v$ by a CDE controlled by $\boldsymbol{X}$ of the form of \eqref{eq:cde}, in which we approximate the unknown vector field $\boldsymbol{f}$ with a neural network, denoted by $\boldsymbol{f}_\theta \colon \reals^v \rightarrow \reals^{v \times d}$. By letting $\zeta \in \reals^v$ be the initial condition, we say that the solution $\boldsymbol{Z}_t \in \reals^v$ to
\begin{equation}
    \dd \boldsymbol{Z}_t = \boldsymbol{f}_\theta (\boldsymbol{Z}_t) \dd \boldsymbol{X}_t, \quad \boldsymbol{Z}_{t_0} = \zeta,
    \label{eq:ncde}
\end{equation}
is the solution of a neural controlled differential equation. 

The difference of a neural CDE compared with the neural ODE of \citet{chen2018neural} is simply that there is a $\dd \boldsymbol{X}_t$ rather than a $\dd t$ term. This modification is what enables the model to change continuously based on incoming data, since the path $\boldsymbol{X}_t$ adapts the dynamics of the system through the $\dd \boldsymbol{X}_t$ term.

Our goal now is to learn $\boldsymbol{f}_\theta$ from labelled training data such that the resulting CDE best models our problem of interest. However, it is not currently clear how this updating can be done, since we do now know how to update $\boldsymbol{Z}_t$ given $\boldsymbol{X}_t$. For this we will need to appeal to existing tools from rough path theory.

\subsubsection*{Training a neural CDE.}
First of all, for any data stream $\boldsymbol{X}$, we need to be able to compute the corresponding hidden state $\boldsymbol{Z}$, that is, to solve the neural CDE \eqref{eq:ncde}. If we instead had a neural ODE, where $\dd \boldsymbol{X}_t$ is $ \dd t$, then the method of solution is straightforward. We integrate the equation resulting in 
\begin{equation*}
    \boldsymbol{Z}_t = \boldsymbol{Z}_{t_0} + \int^t_{t_0} \boldsymbol{f}_\theta (\boldsymbol{Z}_u) \dd u
    \label{eq:node_integral}
\end{equation*}
and update $\boldsymbol{Z}_t$ using a standard integral solver. The integral form of the Neural CDE however is
\begin{equation*}
    \boldsymbol{Z}_t = \boldsymbol{Z}_{t_0} + \int^t_{t_0} \boldsymbol{f}_\theta (\boldsymbol{Z}_u) \dd \boldsymbol{X}_u.
    \label{eq:ncde_integral}
\end{equation*}

Our difficulty now lies in the fact that we need to integrate with respect to the multi-dimensional path control $\boldsymbol{X}_t$. Unsurprisingly, rough path theory holds the solution, and this solution is related to the log-signature of the path and more specifically to the log-ODE method discussed in Section \ref{sec:signatures}.


The log-ODE method describes how the solution of any CDE can be approximated by an ODE that incorporates the log-signature of the path. More precisely, it states that over an interval  $[s, t)$,
\begin{equation*}
    \boldsymbol{Z}_t \approx \boldsymbol{Z}_s + \int_s^t \widehat{\boldsymbol{f}}_\theta(\boldsymbol{Z}_u) \frac{\logsig^n_{s, t}(\boldsymbol{X})}{t - s} \dd u,
\end{equation*}
where the vector field $\widehat{\boldsymbol f}_\theta : \mathbb{R}^v\rightarrow\mathbb{R}^v$ is defined for any $\boldsymbol y = (y_1,...,y_n) \in T^n(\mathbb{R}^d)$ as
\[\widehat{\boldsymbol{f}}_\theta(\boldsymbol{Z}_u) \boldsymbol y = \sum_{k=1}^n \boldsymbol f_\theta^k(\boldsymbol{Z}_u)y_k\]
where for any $k=1,...,n$ the functions $\boldsymbol f^k_\theta : \mathbb{R}^v \to L((\mathbb{R}^d)^{\otimes k}, \mathbb{R}^v)$ are defined recursively by
\begin{align*}
    \boldsymbol f^1_\theta = \boldsymbol f_\theta, \dots , \boldsymbol f^{(k+1)}_\theta = D(\boldsymbol f^k_\theta)\boldsymbol f_\theta,
\end{align*}
where $D(\cdot)$ denotes the Fr\'echet derivative. This allows us to evaluate the Neural CDE model via computation of the log-signature terms followed by application of ODE solvers. 

To train the neural network, that is, to find the ``best'' set of parameters $\theta$ of the neural network $f_\theta$, we therefore proceed as follows. We pick points $r_i$ such that $t_0 < r_0 < \cdots < r_n = t_n$, where the steps may potentially be larger than the discretisation of the data, and update
\begin{align*}
    \boldsymbol{Z}_{r_{i+1}} &= \boldsymbol{Z}_{r_i} \\
    & \quad + \int^{r_{i+1}}_{r_i} \widehat{\boldsymbol{f}}_\theta(u) \frac{\logsig^N_{r_i, r_{i+1}}(\boldsymbol{X})}{r_{i+1} - r_i} \dd u.
\end{align*}
We then employ standard ODE solvers to update $\boldsymbol{Z}_t$, compare the final value against the true labels, and backpropagate the information to update $\boldsymbol{f}_\theta$ \citep{ chen2018neural}. 

\begin{figure}[H]
    \centering
    \includegraphics[width=\linewidth]{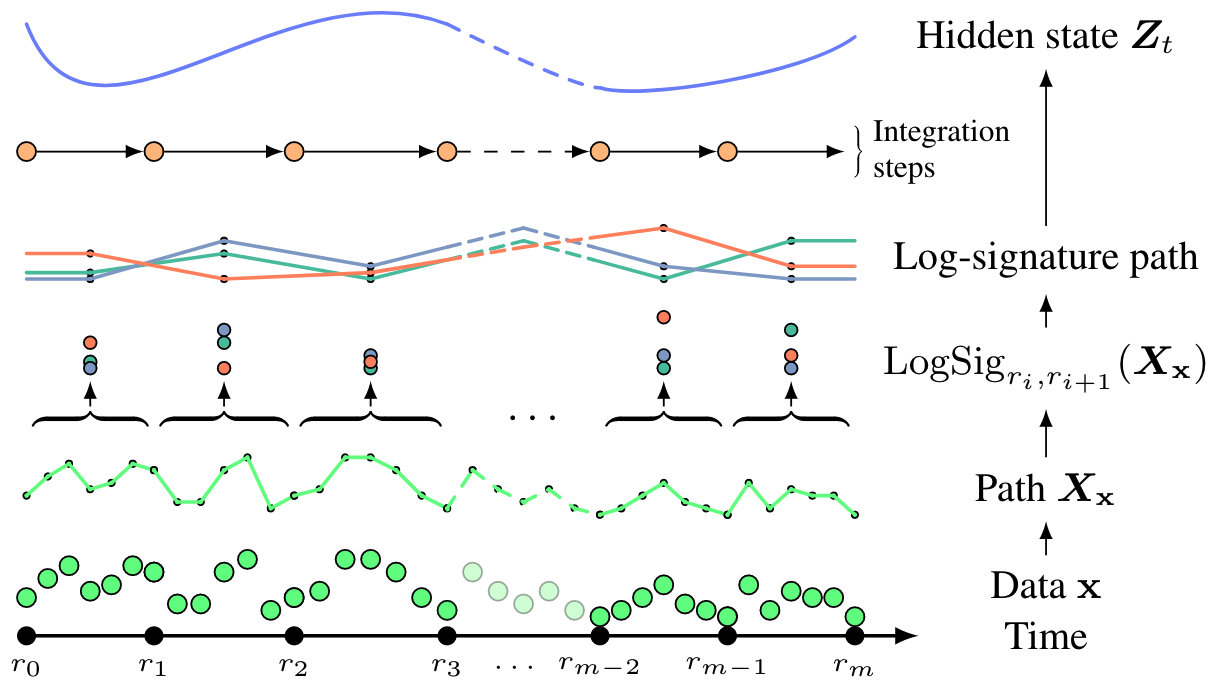}
    \caption{Pictorial description of the log-ode method used to solve a neural CDE. The data is converted onto a continuous embedding which is in turn converted into a collection of log-signatures that are used to update the hidden state.}
    \label{fig:log-ode}
\end{figure}

We give in Figure \ref{fig:log-ode} a graphical depiction of this method.

\subsubsection*{Performance.}

Neural CDEs have been shown to be highly performant across a range of benchmarks. \citet{kidger2020neural} show that the neural CDE outperforms similar ODE-benchmarks on a range of tasks on both performance metrics and memory usage. \citet{morrill2021neural} show that, by updating using higher degree log-signatures and longer steps, improvements in memory usage, training time, and model performance can all be achieved---this was verified on problems of length over 17,000. \citet{bellot2021policy} achieve state-of-the-art results using neural CDEs on counterfactual estimation tasks. A particular type of Neural CDEs are ones driven by Brownian motion, dubbed Neural SDEs. These models can serve as natural generative models for time series \citep{arribas2020sig, kidger2021neural}.

The examples considered were notably far reaching, ranging from the prediction of sepsis in ICUs to understanding the effect of Eurozone membership on the stream of current account deficits. They have also been successfully applied to datasets that are regular or irregular, fully or partially observed, and short problems to problems of long length (length $> 1000$). The ability to apply to problems of long length is facilitated by the rough log-ode update method that enables to trade length for additional log-signature terms making the method extremely useful for long-time series \citep{morrill2021neural}.

To conclude, neural CDEs are a powerful tool for learning streams. In particular, they apply naturally to irregularly sampled and partially observed data. Indeed, the neural CDE interacts with the discrete data $\boldsymbol{x}$ through a continuous time embedding $\boldsymbol{X}$. This means that all data is placed on the same footing, be it irregular, partially observed, or both. This makes it a natural model for handling such non-regular data, which has posed a problem for previous models such as the RNN which expect data that is both regularly spaced and fully observed.

Moreover, the neural CDE retains all the benefits of ODE modelling (adjoint method, continuously defined, error monitoring, ...). This is not true of many other similar models; for example, the ODE-RNN cannot operate using the memory efficient adjoint method of training  \citep[see][for more details]{kidger2020neural}.

%% file: sections/rnns.tex
\subsection{Connection to RNNs}\label{sec:rnns}

Not only the connection between CDEs and neural networks allow to define new high-performance algorithms, but it also allows to better understand classical neural networks. In this subsection, we show that Recurrent Neural Networks (RNN), which are  very popular neural architectures for learning with  sequential data, may be reformulated in the language of rough path theory. This viewpoint is very rich; in particular, it allows to leverage the theory of CDEs and signatures to provide theoretical results on RNN, which is otherwise a very difficult endeavour. 


We place ourselves in a supervised learning setting. The input data is a sample of $N$ i.i.d.~streams $\{\boldsymbol{x}^{(1)}, \dots,\boldsymbol{x}^{(N)}\}$, where each $\boldsymbol{x}^{(i)}$ is a stream as defined by \eqref{eq:def_data_sequence}. However, contrary to the previous section, we assume that all $\boldsymbol{x}^{(i)}$s are sampled on a regular partition: for any $1 \leq j \leq T$, we have $t_j = \nicefrac{j}{T}$. The outputs of the learning problem can be either labels (classification setting) or sequences (sequence-to-sequence setting).


A (residual) recurrent neural network is defined by a sequence of hidden states $z_{1}, \dots, z_{T} \in\reals^v$, where, for $\boldsymbol{x}= (x_1, \dots, x_T)$ a generic data sample, for any $0 \leq j \leq T-1$,
\begin{equation} \label{eq:residual-rnn}
    z_{j+1} = z_{j} + \frac{1}{T}  f_\theta(z_{j}, x_{j+1}).
\end{equation}
At each time step $j$, the output of the network is $y_j = \psi(z_j)$, where $\psi$ is a linear function. The simplest choice for the function $f_\theta$ is the feedforward model (also known as Vanilla RNN), defined by
\begin{equation*}
    f_{\theta}(h, x) = \sigma(U h + V x + b),
\end{equation*}
where $\sigma$ is a nonlinear function applied element-wise (such as sigmoid or hyperbolic tangent), $U \in \reals^{v \times v}$ and $V \in \reals^{v \times d}$ are weight matrices, and $b \in \reals^{v}$ is the bias ($\theta$ is then a vector containing all the coefficients in $U$, $V$, and $b$). Note that the GRU \citep{Cho2014LearningPR} and LSTM \citep{hochreiter1997long} models can also be rewritten under the form \eqref{eq:residual-rnn}.

We let $\boldsymbol{X}: [0,1] \to \reals^d$ denote the embedding of $\boldsymbol{x}$ (such that we have $\boldsymbol{x}_j = \boldsymbol{X}_{\nicefrac{j}{T}}$, $1 \leq j \leq T$). Following the neural ODE paradigm of \citet{chen2018neural}, the recursive equation \eqref{eq:residual-rnn} may be seen as an Euler discretization of the ODE
\begin{equation*}
    \dd \boldsymbol{Z}_t = f_\theta(\boldsymbol{Z}_t, \boldsymbol{X}_t)\dd t, \quad Z_0 =z_0, 
\end{equation*}
where $\boldsymbol{Z}$ is a continuous-time hidden state and $t \in [0,1]$. The output of the network at time $j$ is then $\psi(\boldsymbol{Z}_{\nicefrac{j}{T}})$. As mentioned before, by increasing the dimension of $\boldsymbol{Z}$, any ODE can be rewritten as a CDE of the form
\begin{equation*} \label{eq:continuous-rnn-cde}
    \dd \boldsymbol{Z}_t = \boldsymbol{f}_\theta(\boldsymbol{Z}_t) \dd \boldsymbol{X}_t.
\end{equation*}

An important property of the tensor algebra $T((\reals^d))$ introduced in Section \ref{sec:streams}, is that it can be equipped with a Hilbert structure \citep{Kiraly2016,salvi2021signature}, that is, with a scalar product denoted by $\langle \cdot, \cdot \rangle$. The following real-valued function
\begin{align*}
    \kappa: (\boldsymbol{X}, \boldsymbol{Y}) & \mapsto \langle \sig(\boldsymbol{X}), \sig(\boldsymbol{Y}) \rangle,
\end{align*}
is then well-defined, positive-definite, and called a kernel. The general idea of kernel methods is to map data to a high dimensional space so that linear methods may be used in this space, while being highly nonlinear in the original data \citep[see, e.g., ][]{scholkopf2018learning}. We refer to Section \ref{sec:kernels} for more details on kernel methods with signatures. Using a Taylor expansion of $\boldsymbol{Z}$, under some assumptions on the regularity of $\boldsymbol{f}_{\theta}$, it is then possible to show \citep[][Theorem 1]{fermanian2021framing} that there exists $\alpha(\theta) \in T((\reals^d))$ such that
\begin{equation*}   
    |y_T - \langle \alpha(\theta),\sig(\boldsymbol{X})  \rangle| \leq \frac{C}{T},
\end{equation*}
where $\alpha(\theta)$ depends only on the parameters of the RNN, $C$ is a constant, and $y_T$ is the final output of the RNN. In other words, if $g_\theta$ is the function that maps an input sequence $\boldsymbol{x}$ to the final output of the RNN (that is, $g_\theta(\boldsymbol{x}) = \psi(z_T)$), and if $\mathcal{H}$ is the reproducing kernel Hilbert space (RKHS) associated to $\kappa$, then $g_\theta \in \mathcal{H}$ at a $\mathcal{O}(\nicefrac{1}{T})$ error term.

This result allows to reinterpret the action of the recurrent network (RNN) as a scalar product in an (infinite-dimensional) Hilbert space, thereby framing the RNN as a kernel method. The key ingredients to obtain this Hilbert space are the Neural CDE model and the signature kernel.

A first consequence is that it gives natural generalization bounds under mild assumptions. For example, assume a binary classification problem, that is, we are given an i.i.d.~sample of $N$ random pairs of observations $(\boldsymbol{x}^{(i)}, \boldsymbol{y}^{(i)})$, where $\boldsymbol{y}^{(i)} \in \{-1, 1\}$.  In this context, the network outputs a real number $g_{\theta}(\boldsymbol{x})=\psi(z_T) \in \reals$ and the predicted class is $2 \cdot \boldsymbol{1}(g_{\theta}(\boldsymbol{x}) > 0) - 1$. The parameters $\theta$ of the RNN are fitted by empirical risk minimization using a loss function $\ell:\reals \times \reals \to \reals^{+}$, for example the logistic loss. The training loss is defined by
\begin{equation*}
   \widehat{\mathcal{R}}_{N}(\theta) = \frac{1}{N} \sum_{i=1}^N \ell \big(\boldsymbol{y}^{(i)}, g_{\theta}(\boldsymbol{x}^{(i)}) \big),
\end{equation*}
 and we let $\widehat{\theta}_N \in \textnormal{argmin }_{\theta \in \Theta } \widehat{\mathcal{R}}_N (\theta)$. Using the kernel approach described above, \citet[][Theorem 2]{fermanian2021framing} obtain the following informal result. Under some assumptions on its weight matrices, for a feedforward RNN, there exists a constant $B>0$ such that for any $ \theta$, $\| \alpha(\theta) \| \leq B$. Then with probability at least $1-\delta$,
    \begin{align*} 
        &\mathbb{P}\big(\boldsymbol{y}g_{\widehat{\theta}_N}(\boldsymbol{x}) \leq 0| \mathcal{D}_N \big) \\
        & \quad \leq \widehat{\mathcal{R}}_N(\widehat{\theta}_N) + \frac{c_2}{T} + \frac{8 B K_{\ell} }{(1-L)\sqrt{N}} \\
        & \qquad + \frac{2BK_{\ell}}{1-L} \sqrt{\frac{\log(\nicefrac{1}{\delta})}{2N}}.
    \end{align*}

The term in $\mathcal{O}(\nicefrac{1}{T})$ comes from the continuous-time approximation, whereas the speed in $\nicefrac{1}{\sqrt{N}}$ is classical: the detour by CDEs provides a theoretical framework adapted to RNN, at the modest price of an additional $\mathcal{O}(\nicefrac{1}{T})$ term. 

In addition to providing a sound theoretical framework, framing deep learning in an RKHS provides a natural norm, which can be used for regularization. In our context, for two streams $\boldsymbol{x}$ and $\boldsymbol{y}$, we can bound the difference between the RNN outputs $g_\theta(\boldsymbol{x})$ and $g_\theta(\boldsymbol{y})$ by using the Cauchy-Schwartz inequality, as follows:
\begin{align*}
&|g_\theta(\boldsymbol{x})-g_\theta(\boldsymbol{y})| \\
& \quad \leq \frac{C}{T} + |\langle \alpha(\theta),S(\boldsymbol{X}) - S(\boldsymbol{Y})  \rangle| \\
& \quad \leq  \frac{C}{T} + \|S(\boldsymbol{X}) - S(\boldsymbol{Y}) \| \|\alpha(\theta)\|.
\end{align*}
If $\boldsymbol{x}$ and $\boldsymbol{y}$ are close, so are their embeddings $\boldsymbol{X}$ and $\boldsymbol{Y}$, and the term $\|S(\boldsymbol{X}) - S(\boldsymbol{Y}) \|$ is therefore small \citep[][Proposition 7.66]{friz2010multidimensional}. When $T$ is large, we see that the magnitude of $ \|\alpha(\theta)\|$ determines how close the predictions are. A natural training strategy to ensure stable predictions is then to penalize the problem by minimizing the loss $\widehat{\mathcal{R}}_n(\theta) + \lambda  \|\alpha(\theta)\|$.



To conclude, framing RNN as discretizations of CDEs is a powerful theoretical tool. It shows that the signature kernel appears naturally when analysing RNN. This kernel can also be used on its own to define new algorithms for data streams, which we now discuss.

%% file: sections/kernels.tex
\section{Signature kernels} \label{sec:kernels}

Kernel methods have been shown to be efficient learning techniques in situations where inputs are non-Euclidean and the number of training instances is limited so that deep-learning-based techniques cannot be easily deployed \citep[see][for a detailed account on kernel methods in machine learning]{hofmann2008kernel}.  When the input data is sequential, the design of appropriate kernel functions is a notably challenging task. 

Consider a set $\Omega$ and let $\phi:\Omega\rightarrow H$ be a \emph{feature map} embedding elements of $\Omega$ into a Hilbert space $H$ (referred to as the \emph{feature space}).

For most of the popular kernels (RBF, Mat\'ern ...) $H$ is infinite dimensional and $\phi$ is either hard to compute or doesn't admit an explicit expression. Nonetheless, the kernel $\kappa$ can often be evaluated efficiently at any pair of points $x,y \in \Omega$ without referring back to the feature map. This property is known as  \emph{kernel trick}. 

Given a pair of $\Omega$-valued paths $\boldsymbol{X},\boldsymbol{Y}$, one can lift them both to two $H$-valued paths $\mathbb{X},\mathbb{Y}$ by means of the feature map $\phi$ as
\begin{equation*}
    \mathbb{X}_s = \phi(\boldsymbol{X}_s), \quad \mathbb{Y}_t=\phi(\boldsymbol{Y}_t).
\end{equation*}
Under the assumption that $\mathbb{X}$ and $\mathbb{Y}$ are continuous and have bounded variation, their signatures $\sig(\mathbb{X}), \sig(\mathbb{Y})$ are well defined as elements of $T((V))$. 
Hence, the signature allows to turn a kernel $\kappa$ on $\Omega$ into a kernel $\kappa_{\text{sig}}$ on $\Omega$-valued paths defined as follows
\begin{equation*}
    \kappa_{\text{sig}}(\boldsymbol{X}_{s,s'},\boldsymbol{Y}_{t,t'}) = \left(\sig_{s,s'}(\mathbb{X}), \sig_{t,t'}(\mathbb{Y})\right).
\end{equation*}
We call $\kappa_{\text{sig}}$ the \emph{signature kernel} associated to the kernel $\kappa$. A natural question at this stage is: is there a kernel trick for $\kappa_{\text{sig}}$? 

The answer is positive and given by \citet[][Theorem 2.5]{salvi2021signature} who make a link between signature kernels and PDEs. In the simple case where the paths $\boldsymbol{X}, \boldsymbol{Y}$ are differentiable, the theorem states that $\kappa_{\text{sig}}(\boldsymbol{X}_{s,s'},\boldsymbol{Y}_{t,t'}) = U(s',t')$, where $U$ is the solution of the following linear hyperbolic PDE
\begin{equation}\label{eqn:PDE}
    \frac{\partial^2 U}{\partial p \partial q} = \kappa (\dot{\boldsymbol{X}}_p, \dot{\boldsymbol{Y}}_q)U
\end{equation}
with $U(s,\cdot)=U(\cdot,t)=1$, and where $\dot{\boldsymbol{X}}$ stands for the first time derivative of $\boldsymbol{X}$. The PDE \eqref{eqn:PDE} can be solved numerically using the finite differece schemes proposed in \citet{salvi2021signature} and implemented in the library \texttt{sigkernel}\footnote{\url{https://github.com/crispitagorico/sigkernel}}.

The availability of a kernel on paths allows to adapt standard kernel-based machine learning methods to the case of sequential data. It is straightforward to interface kernel estimators from \texttt{scikit-learn} \citep{pedregosa2011scikit} with \texttt{sigkernel} functionalities to build pipelines for time series classification (e.g. Support Vector Machine (SVM)), regression (e.g. Kernel-Ridge) or dimensionality reduction (e.g. kernel PCA) with no more than $10$ lines of code. 

For example, the table below shows classification accuracies (\%) on various UEA time series datasets \citep{bagnall2018uea} obtained by equipping a SVM classifier 
with different choices of kernels \footnote{When the time series are of the same length, standard kernels on $\mathbb{R}^d$ (Linear, RBF) can be used by stacking each channel into one single vector. When the series are not of the same length, other than the signature kernel, the \emph{global alignment kernel} \citep{cuturi2011fast} is a popular choice.}, including the signature kernel. As the results show, the signature kernel (SigKer) is systematically among the top $2$ classifiers across all the datasets (except for FingerMovements and UWaveGestureLibrary) and always outperforms its truncated counterpart (Sig$^N$). 

    \begin{center}
        \resizebox{.48\textwidth}{!}{
        \begin{tabular}{lccccc}
        \toprule
        \textbf{Datasets/Kernels} &
        Linear &
        RBF &
        GAK & 
        Sig$^N$ &
        SigKer \\
        \midrule
        ArticularyWordRecognition & 98.0 & 98.0 & 98.0 & 92.3 & \textbf{98.3}\\
        BasicMotions & 87.5 & 97.5 & 97.5 & 97.5 & \textbf{100.0} \\
        Cricket & 91.7 & 91.7 & \textbf{97.2} & 86.1 & \textbf{97.2}\\
        ERing & 92.2 & 92.2 & \textbf{93.7} & 84.1 & 93.3\\
        Libras & 73.9 & 77.2 & 79.0 & \textbf{81.7} & \textbf{81.7}\\
        NATOPS & 90.0 & 92.2 & 90.6 & 88.3 & \textbf{93.3}\\
        RacketSports & 76.9 & 78.3 & 84.2 & 80.2 & \textbf{84.9}\\
        FingerMovements & 57.0 & 60.0 & \textbf{61.0} & 51.0 & 58.0\\
        Heartbeat & 70.2 & 73.2 & 70.2 & 72.2 & \textbf{73.6}\\
        SelfRegulationSCP1 & 86.7 & 87.3 & \textbf{92.4} & 75.4 & 88.7\\
        UWaveGestureLibrary & 80.0 & \textbf{87.5} & \textbf{87.5} & 83.4 & 87.0\\
        \bottomrule
        \end{tabular}
    }
    \end{center}

If a kernel is \emph{characteristic} \citep[see][for more details]{gretton2012kernel} and operates on the common support of two random variables, it can be used to define a notion of distance between random variables from finite collections of samples. This distance is commonly referred to as \emph{maximum mean discrepancy (MMD)} and can be used in practice to quantify the (dis)similarity between two datasets, hypothesis testing etc. 

\begin{figure*}[ht]
\centering
\includegraphics[width=0.8\textwidth,trim={0 4.1cm 0 0},clip]{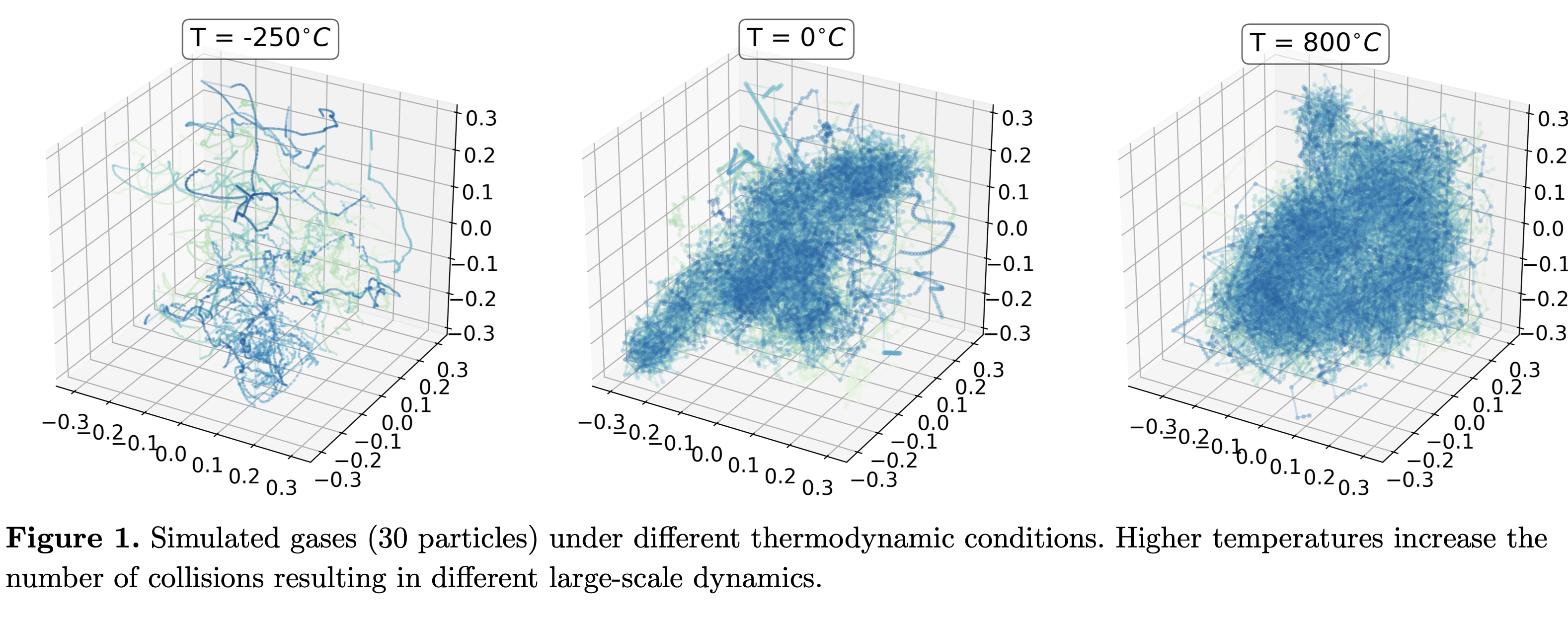}
\caption{\small Simulation of the trajectories traced by $20$ particles of an ideal gas in a $3$-d box under different thermodynamic conditions. Higher temperatures equate to a higher internal energy in the system which increases the number of collisions resulting in different large-scale dynamics of the gas.}
\label{fig:ideal_gas}
\end{figure*}

When the random variables are supported on path-space, they are called stochastic processes. Because the signature kernel is characteristic \citep[][Theorem 5.3]{chevyrev2018signature}, it allows to define a MMD on stochastic processes. Indeed, for any two compactly supported probability measures $\mathbb{P},\mathbb{Q}$ on path-space, one has the following representation of the MMD \citep[][Lemma 6]{gretton2012kernel}
\begin{align*}
    d_{\text{MMD}}(\mathbb{P}, \mathbb{Q})^2 &= \mathbb{E}_{(\boldsymbol{X},\boldsymbol{X}') \sim \mathbb{P} \times \mathbb{P}}[\kappa_{\text{sig}}(\boldsymbol{X},\boldsymbol{X}')] \\
    &- 2\mathbb{E}_{(\boldsymbol{X},\boldsymbol{Y})\sim \mathbb{P} \times \mathbb{Q}}[\kappa_{\text{sig}}(\boldsymbol{X},\boldsymbol{Y})]\\
    &+ \mathbb{E}_{(\boldsymbol{Y},\boldsymbol{Y}') \sim \mathbb{Q} \times \mathbb{Q}}[\kappa_{\text{sig}}(\boldsymbol{Y},\boldsymbol{Y}')].
\end{align*}
Given $M$ sample paths $\{\boldsymbol{X}^i\}_{i=1}^M \sim \mathbb{P}$ and $N$ sample paths $\{\boldsymbol{Y}^i\}_{i=1}^N \sim \mathbb{Q}$, the MMD can be computed in practice via the following unbiased estimator for $d_{\text{MMD}}(\mathbb{P}, \mathbb{Q})^2:$
\begin{align*}
    &\frac{1}{M(M-1)}  \sum_{i=1}^M\sum_{j\neq i}^M \kappa_{\text{sig}}(\boldsymbol{X}^i,\boldsymbol{X}^j) \\
    &- \frac{2}{MN}\sum_{i=1}^M\sum_{j=1}^N \kappa_{\text{sig}}(\boldsymbol{X}^i,\boldsymbol{Y}^j) \\
    &+ \frac{1}{N(N-1)}\sum_{i=1}^N\sum_{j\neq i}^N \kappa_{\text{sig}}(\boldsymbol{Y}^i,\boldsymbol{Y}^j).
\end{align*}
The characteristicness of $\kappa_{\text{sig}}$ makes the corresponding MMD an efficient statistics to distinguish two stochastic processes \citep[][Theorem 5]{gretton2012kernel}
\begin{equation*}
    d_{\text{MMD}}(\mathbb{P}, \mathbb{Q}) = 0 \iff \mathbb{P} = \mathbb{Q},
\end{equation*}
and consequently allows to build two-sample hypothesis tests for the null hypothesis
$H_0 : \mathbb{P} = \mathbb{Q}$ against the alternative $H_1: \mathbb{P} \neq \mathbb{Q}$.
The signature kernel has been applied in a variety of kernel-based contexts dealing with sequential data such as Gaussian Processes \citep{toth2020bayesian, lemercier2021siggpde} or Distribution Regression (DR) \citep{lemercier2021distribution, salvi2021higher}.

For example, in the context of DR, \citet{lemercier2021distribution} make use of the signature kernel to recover some thermodynamic properties of an ideal gas (e.g. the temperature) from the individual trajectories of its particles (see \Cref{fig:ideal_gas}). The complexity of the global dynamics of the gas depends on the environmental temperature ($T$) as well as on the radius of the particles ($r$). For a fixed temperature $T$, the larger the radius the higher the chance of collision between the particles. In the table below we report the DR results (MSE $10^{-2}$) of two experiments on simulated data: one where particles have a small radius (few collisions) and another where they have a bigger radius (many collisions). The performance of the signature kernel are superior but comparable to the ones produced by other models in the simpler setting (few collisions). However, in the presence of a high number of collisions, the signature kernel becomes more informative to retrieve the global temperature from local trajectories, while the performance from all other kernels deteriorates \citep[for additional experimental details see][]{lemercier2021distribution}.

\begin{center}
    \resizebox{.48\textwidth}{!}{
    \begin{tabular}{lcc}
    \toprule
    \textbf{Kernels/Regime} &
    Few collisions & Many collisions\\
    \midrule
    RBF & 3.08~(0.39) &  4.36~(0.64) \\
    Mat\'ern32 & 3.54~(0.48) &  4.12~(0.39)\\
    GAK & 2.85~(0.43) &  3.69~(0.36) \\
    SigKer &\textbf{1.31}~(0.34) &  \textbf{0.08}~(0.02)\\
    \bottomrule
    \end{tabular}
    }
\end{center}

%% file: sections/conclusion.tex
\section{Conclusion}

This paper attempts to provide a concise overview of some of the recent advances in the application of rough path theory to machine learning. Controlled differential equations (CDEs) are discussed as the key mathematical model to describe the interaction of a stream with a physical control system. The signature naturally arises in the description of the response produced by such interactions. The signature comes equipped with a variety of powerful properties rendering it an ideal feature map for streamed data. Next, we summarised recent advances in the symbiosis between deep learning and CDEs, studying the link with RNNs and culminating with the Neural CDE model. We concluded with a discussion on signature kernel methods, which are an alternative to deep learning methods.

\section*{Acknowledgements}

Terry Lyons’s contribution to this work is supported by the EPSRC program grant DataSig [grant number EP/S026347/1], and its project partners: in part by The Alan Turing Institute under the EPSRC grant EP/N510129/1, in part by The Alan Turing Institute’s Data Centric Engineering Programme under the Lloyd’s Register Foundation grant G0095, in part by The Alan Turing Institute’s Defence and Security Programme, funded by the UK Government, and in part by the Hong Kong Innovation and Technology Commission (InnoHK Project CIMDA).